\documentclass{article}
\usepackage{stywhispers,amsmath,epsfig}

\usepackage[per-mode=symbol]{siunitx}
\usepackage{upgreek}
\usepackage[hidelinks]{hyperref}
\usepackage{tabularx}
\usepackage{booktabs}
\usepackage{subfig}
\usepackage{multirow}
\usepackage{color}
\usepackage{paralist}
\usepackage[super]{nth}
\usepackage[nameinlink, capitalize]{cleveref}
\usepackage{float}

\title{Machine learning regression on hyperspectral data to estimate multiple water parameters}
\name{Philipp M. Maier, Sina Keller \thanks{Thanks to the German Federal Ministry of Education and Research for funding the WAQUAVID project.}}
\address{Karlsruhe~Institute~of~Technology~(KIT)\\
	Institute~of~Photogrammetry~and~Remote~Sensing~(IPF),\\
	Englerstra{\ss}e 7, D-76131 Karlsruhe, Germany}

\begin{document}
%
\maketitle
\vspace{-1.5cm}
\noindent\copyright\ 2018 IEEE
\begin{abstract}
In this paper, we present a regression framework involving several machine learning models to estimate water parameters based on hyperspectral data. 
Measurements from a multi-sensor field campaign, conducted on the River Elbe, Germany, represent the benchmark dataset. 
It contains hyperspectral data and the five water parameters chlorophyll~\textit{a}, green algae, diatoms, CDOM and turbidity. 
We apply a PCA for the high-dimensional data as a possible preprocessing step.
Then, we evaluate the performance of the regression framework with and without this preprocessing step. 
The regression results of the framework clearly reveal the potential of estimating water parameters based on hyperspectral data with machine learning.
The proposed framework provides the basis for further investigations, such as adapting the framework to estimate water parameters of different inland waters.
\end{abstract}


\section{Introduction}
\label{sec:intro}

Monitoring the quality of inland waters is a major research topic in many scopes of environmental applications.
Traditionally, a water body is monitored with several point-based in-situ measurements.
Although, the data measured at such specific points is precise, these measurement techniques are time consuming when covering a large area. Additionally, if area-wide information is required, the data needs to be interpolated as well.
Regarding the demand for area-wide information of water parameters, hyperspectral sensing seems to close the gap left behind by in-situ point-measurements.

The spectral signature of natural water bodies is mainly characterized by absorption and scattering on four water parameters: chlorophyll~\textit{a}, suspended solids, colored dissolved organic matter (CDOM) and turbidity~\cite{Gitelson.1990,HUNTER.2008}. 

Chlorophyll~\textit{a} is a pigment which functions as an indicator for algae existence and nutrition supply in water bodies.
Turbidity is influenced by algae concentrations and suspended particles.
CDOM is a natural degradation product group of organic materials which is natural in waters.  
It is also affected by human impacts like sewage treatment plants.

Recent approaches were made to estimate chlorophyll~\textit{a} concentration, CDOM content and turbidity with hyperspectral data.
As described in~\cite{Gitelson.1990}, the spectral signature of chlorophyll~\textit{a} in water is characterized by a reflectance minimum at about \SI{670}{\nano\meter} as well as reflectance maxima in the green and red spectral range.
The distinction of algae species based on their measured hyperspectral signature was studied e.g. in a laboratory~\cite{HUNTER.2008}.
With increasing CDOM content, the absorption simultaneously increases in shorter wavelengths~\cite{Menken.2006}. 

Estimations of water parameters with hyperspectral data mostly rely on empirical approaches.
Centerpiece of these approaches is, on the one hand, the selection of reflectance bands including single bands, band ratios and multiple band algorithms~\cite{Menken.2006, GITELSON.1992, Rundquist.1996, Fraser.1998, Koponen.2002, Bhatti.2010, Matthews.2010, Zhou.2013, Kisevic.2016}.
On the other hand, further studies consider the area  between \SIrange{672}{742}{\nano\meter} under the related maximum~\cite{Mannheim.2004,Schalles.1998} or the maximum itself~\cite{Gitelson.1990,Mannheim.2004,Schalles.1998} for the estimation of chlorophyll~\textit{a} concentration. 
In addition, derivatives of the spectra up to the fourth grade are applied~\cite{Rundquist.1996,Kisevic.2016}.

\begin{figure*}[tb]
	\centering
    \includegraphics[width=0.95\textwidth]{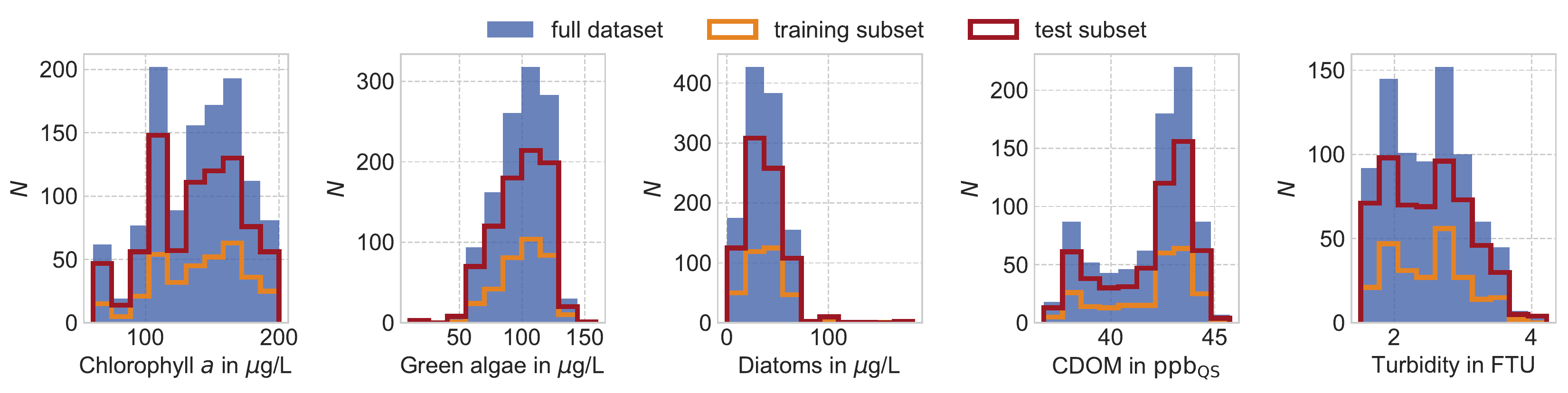}
    \caption{Distributions of the regression target variables (water parameters) in the training (orange line) and the test (red line) dataset. $N$ corresponds to the number of datapoints.\label{fig:datahist_all}}
    \vspace{-0.5cm}
\end{figure*}

Machine learning approaches to estimate CDOM and \mbox{chlorophyll~\textit{a}} are applied yet solely in few studies. 
The former is estimated with e.g. functional linear models~\cite{Yu.2010} or linear stepwise regression~\cite{Brezonik.2015}.
Currently, random forests are applied to estimate the latter~\cite{Maier.2018}. 
In general, machine learning models require sufficient input data to solve a non-linear regression problem with high-dimensional data.

In this paper, we proceed to examine the potential of supervised machine learning approaches to estimate crucial water parameters such as chlorophyll~\textit{a}, green algae, diatoms, CDOM, and turbidity. 
We rely on a dataset which was measured on the German part of the River Elbe under real-world conditions.
Precise monitoring of the water parameters followed by hyperspectral snapshots and in-situ probe measurements were conducted.  
As a remark, a first impression and a description of the dataset is given in~\cite{Maier.2018}.

The major contributions of this paper are: 
\begin{compactitem}
	\item a detailed analysis of the potential of hyperspectral data to estimate these water parameters;
	\item an adequate framework with five regression models: k-nearest neighbor (k-NN), random forest (RF), support vector machine (SVM),  multivariate adaptive regression splines (MARS) and extreme gradient boosting (XGB);
	\item an comprehensive evaluation of the regression performance with and without preprocessing.
\end{compactitem}

We briefly describe the dataset used for the estimation of water parameters based on the introduced regression framework in \cref{sec:data}.
The presentation of the methodology follows in \cref{sec:methods}.
We assess the regression performance of the five machine learning models in \cref{sec:results}.
Finally, we briefly conclude our studies in \cref{sec:conclusion}.

\section{Dataset}
\label{sec:data}

To evaluate the potential of hyperspectral data as input data for the estimation of water parameters, we rely on a dataset which was sampled during a field campaign in summer 2017 on the River Elbe.
A hyperspectral sensor Cubert UHD 285 was mounted on a research ship.
The PhycoSens sensor, an invention of BBE Moldaenke, sampled the concentration of chlorophyll~\textit{a}, green algae and diatoms.
Additionally, the Biofish sensor system~\cite{Holbach.2014} measured the concentration of CDOM and the turbidity.
The distributions of the measured water parameters are illustrated in \cref{fig:datahist_all}.
A detailed description of the actual data acquisition is presented in~\cite{Maier.2018}.

As for the hyperspectral snapshot data, each image contains $50\times 50$ pixels and $125$ spectral channels ranging from \SIrange{450}{950}{\nano\meter} with a spectral resolution of \SI{4}{\nano\meter}.
A selection of the hyperspectral bands is applied to avoid sensor artifacts.
This results in a range of wavelengths from \SIrange{470}{910}{\nano\meter}.
Furthermore, the measured mean spectra of all images were calculated by manually selecting an area, which was undisturbed by bubble formations, shadows or waves.

The sampled reference data of the PhycoSens data is extended by a linear interpolation.  
This interpolation is appropriate, since the measured concentrations change in a continuous matter.
The dataset emerging of the measurements with the PhycoSens counts 1163 high-dimensional datapoints defined by selected hyperspectral bands and the concentrations of chlorophyll~\textit{a}, green algae as well as diatoms as reference data.
Hyperspectral datapoints, which are convenient to the Biofish sensor system data, consist of \si{802} values combining selected hyperspectral bands, CDOM, and turbidity as references.

\section{Methodology}
\label{sec:methods}

To estimate water parameters with hyperspectral data, we use the given dataset for our proposed regression framework.
We apply five supervised machine learning models: k-nearest neighbor (k-NN), random forest (RF), support vector machine (SVM), multivariate adaptive regression splines (MARS) and extreme gradient boosting (XGB).
They perform the regression with the hyperspectral data as input vector and the respective water parameter as target value. 
\cref{tab:implementation} provides a brief summary of all model implementations. 

The datasets (cf. \cref{sec:data}) are split into a training and a test subset in a ratio of $3:7$.  
All models of the framework are trained on the respective training subset and are evaluated on the respective test subset. 
For a reasonable estimation of each water parameter, the distributions of the subsets have to be representative of the specific water parameter's distribution.
The split ratio as well as the distinction between training and test subset decrease overfitting. 
\cref{fig:datahist_all} provides histograms of the distribution of each water parameter. 

To reduce the dimensionality of the hyperspectral input data, we apply a principal component analysis (PCA) as a preprocessing step.
We choose the first eight principal components to fit the models since they cover \SI{99.9}{\percent} of the overall variability.
Alternatively, the regression framework performs the estimation of the water parameters without PCA to which we refer as raw bands in \cref{sec:results}. 
With respect to the k-NN, SVM, and MARS regression models, the input data always is scaled~\cite{altman1992an,vapnik1995the,Friedman1991}.

\section{Results}
\label{sec:results}

\begin{figure*}[htb]
	\centering
    \includegraphics[width=1\textwidth]	{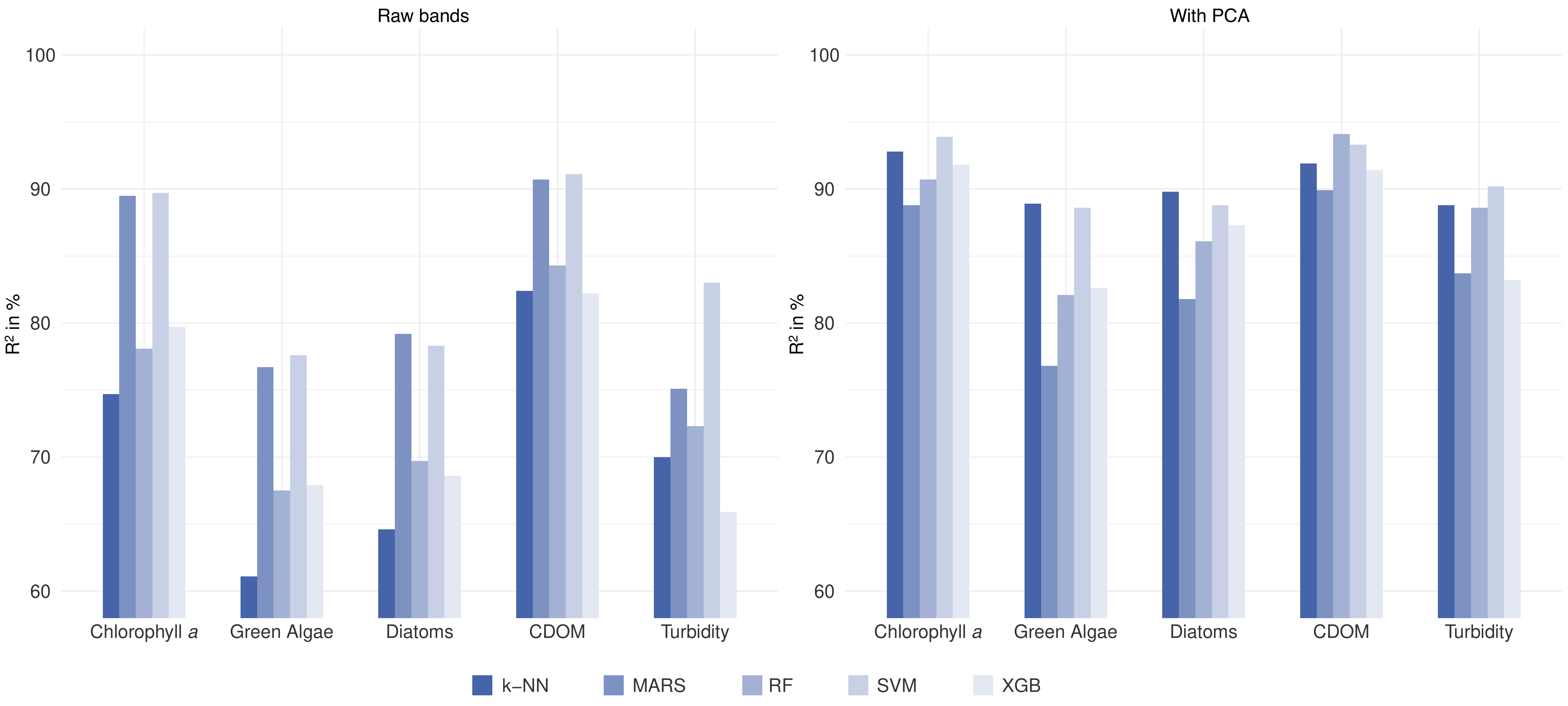}
    \caption{Regression results for the estimation of the water parameters expressed as $R^2$ without PCA (left) and with PCA (right).}
    \label{fig:Figure1}
\end{figure*}

The regression results of the framework for the estimation of the five water parameters are summarized in \cref{fig:Figure1}. 
The regression performance is expressed in terms of the coefficient of determination ($R^2$) and the root mean squared error (RMSE).
The results reveal that the five regression models estimate all water parameters with solely hyperspectral input data remarkably well.
\cref{tab:results_chla,tab:results_greenalgae,tab:results_diatoms,tab:results_cdom,tab:results_turbidity} contain the regression results sorted by water parameters. 
In total, RF, SVM and k-NN models perform better than the MARS and XGB models.

The best estimation performance of each model is obtained by applying the PCA as preprocessing step. 
The chlorophyll~\textit{a} and CDOM concentrations are estimated well by the regression framework (cf. \cref{tab:results_chla,tab:results_diatoms}). 
Each model reaches $R^2$ of approximately \SI{90}{\percent} with PCA.
The estimation performances of diatoms, green algae, and turbidity concentrations achieve $R^2 >$ \SI{80}{\percent} with PCA (cf. \cref{tab:results_greenalgae,tab:results_diatoms,tab:results_turbidity}).

The MARS regression model performs worst compared to the others.
RF provides the best performances of estimating CDOM while the best estimation of the turbidity is achieved by the SVM regression.
The SVM and the k-NN model outperform the other models when estimating the concentrations of chlorophyll~\textit{a}, green algae, and diatoms.
We point out, that outliers in the distribution of the reference data (cf. \cref{fig:datahist_all}) might influence any regression results.

\section{Conclusion}
\label{sec:conclusion}

In this paper, we address the estimation of water parameters with hyperspectral data.
In contrast to most approaches modeling such water parameters, we apply machine learning models.
Our major objective is to evaluate the potential of such models when estimating the amount of chlorophyll~\textit{a}, green algae, diatoms, CDOM or turbidity in inland waters.

We introduce a regression framework involving five selected machine learning models.
As benchmark data, we rely on a dataset, conducted on the River Elbe in Germany which is characterized by measurements of five distinct water parameters.
The regression is conducted with and without PCA to evaluate the impact of dimensionality reduction on the performance due to the applied dimensionality reduction of the input data.
The regression results of the framework clearly reveal the potential of estimating water parameters based on hyperspectral data with machine learning.

The advantages of this framework, in contrast to commonly applied methods, e.g. band ratios, is the ability of machine learning to handle high-dimensional regression problems.
This ability is even given when performing on small, measured datasets.
To conclude our regression framework serves as starting point for further investigations like adapting the framework to estimate water parameters of different types of inland waters.

\begin{table}[tb]
	\centering
	\caption{Implementations of the regression models with the respective R packages.}
	\begin{tabular}{lcl}
	\toprule
	Regression model & Package & Reference\\
	 \midrule
	k-NN 	& caret & \cite{altman1992an}\\
    RF 	 	& ranger & \cite{breiman2001random,geurts2006extremely}\\
    SVM		& e1071	& \cite{vapnik1995the}\\
	MARS	& earth & \cite{Friedman1991} \\
	XGB		& xgboost & \cite{Friedman}\\
	\bottomrule
	\end{tabular}
	\label{tab:implementation}	
\end{table}

\vspace{0.5cm}

\renewcommand{\arraystretch}{1.0}
\begin{table}[h]
	\centering	
    \caption{Regression results of the framework for chlorophyll~\textit{a} estimation.}
	\begin{tabular}{lSSSS}
		\toprule
		\multirow{ 3}{*}{Model} &\multicolumn{2}{c}{raw bands} &\multicolumn{2}{c}{with PCA} \\
		\cmidrule(ll){2-3}
		\cmidrule(ll){4-5}
		 & {$R^2$} & {RMSE}  & {$R^2$} & {RMSE} \\
		 & {in $\%$} & {in \si{\micro\gram\per\liter}} &  {in $\%$} & {in \si{\micro\gram\per\liter}}  \\
		\midrule
        k-NN    & 74.7 & 17.2 & 92.8 & 9.1 \\
		RF      & 78.1 & 16.1 & 90.7 & 11.5 \\
		SVM     & 89.7 & 10.9 & 93.9 & 8.3 \\
		MARS    & 89.5 & 11.2 & 88.8 & 11.4 \\
		XGB     & 79.7 & 15.3 & 91.8 & 9.7  \\
		\bottomrule
	\end{tabular}
	\label{tab:results_chla}~
\end{table}

\vspace{0.5cm}

\renewcommand{\arraystretch}{1.0}
\begin{table}[!h]
	\centering
	\caption{Regression result of the framework for green algae estimation.}
	\begin{tabular}{lSSSS}
		\toprule
		\multirow{ 3}{*}{Model} &\multicolumn{2}{c}{raw bands} &\multicolumn{2}{c}{with PCA} \\
		\cmidrule(ll){2-3}
		\cmidrule(ll){4-5}
		 & {$R^2$} & {RMSE}  & {$R^2$} & {RMSE} \\
		 & {in $\%$} & {in \si{\micro\gram\per\liter}} &  {in $\%$} & {in \si{\micro\gram\per\liter}}  \\
		\midrule
        k-NN    & 61.1 & 13.2 & 88.9 & 7.1 \\
		RF      & 67.5 & 12.2 & 82.1 & 9.4 \\
		SVM     & 77.6 & 10.0 & 88.6 & 7.2 \\
		MARS    & 76.7 & 10.2 & 76.8 & 10.2 \\
		XGB     & 67.9 & 12.0 & 82.6 & 8.8  \\
		\bottomrule
	\end{tabular}
	\label{tab:results_greenalgae}
\end{table}

\vspace{0.5cm}

\renewcommand{\arraystretch}{1.0}
\begin{table}[!h]
	\centering
	\caption{Regression result of the framework for diatoms estimation.}
	\begin{tabular}{lSSSS}
		\toprule
		\multirow{ 3}{*}{Model} &\multicolumn{2}{c}{raw bands} &\multicolumn{2}{c}{with PCA} \\
		\cmidrule(ll){2-3}
		\cmidrule(ll){4-5}
		 & {$R^2$} & {RMSE}  & {$R^2$} & {RMSE} \\
		 & {in $\%$} & {in \si{\micro\gram\per\liter}} &  {in $\%$} & {in \si{\micro\gram\per\liter}}  \\
		\midrule
        k-NN    & 64.6 & 11.5 & 89.8 & 6.2 \\
		RF      & 69.7 & 10.9 & 86.1 & 8.1 \\
		SVM     & 78.3 & 9.0  & 88.8 & 6.5 \\
		MARS    & 79.2 & 8.9  & 81.8 & 8.2 \\
		XGB     & 68.6 & 10.8 & 87.3 & 7.0  \\
		\bottomrule
	\end{tabular}
	\label{tab:results_diatoms}
\end{table}

\renewcommand{\arraystretch}{1.0}
\begin{table}[!h]
	\centering
	\caption{Regression result of the framework for CDOM estimation. CDOM is measured in parts per billion, ${\mathrm{ppb}_{\mathrm{QS}} = 10^{-9}}$ and is calibrated against Quinine Sulfate (QS).}
	\begin{tabular}{lSSSS}
		\toprule
		\multirow{ 3}{*}{Model} &\multicolumn{2}{c}{raw bands} &\multicolumn{2}{c}{with PCA} \\
		\cmidrule(ll){2-3}
		\cmidrule(ll){4-5}
		 & {$R^2$} & {RMSE}  & {$R^2$} & {RMSE} \\
		 & {in $\%$} & {in $\mathrm{ppb}_{\mathrm{QS}}$} &  {in $\%$} &  {in $\mathrm{ppb}_{\mathrm{QS}}$}  \\
		\midrule
        k-NN    & 82.4 & 0.9 & 91.9 & 0.6 \\
		RF      & 84.3 & 0.8 & 94.1 & 0.5 \\
		SVM     & 91.1 & 0.6 & 93.3 & 0.5 \\
		MARS    & 90.7 & 0.6 & 89.9 & 0.7 \\
		XGB     & 82.2 & 0.9 & 91.4 & 0.6  \\
		\bottomrule
	\end{tabular}
	\label{tab:results_cdom}
\end{table}

\renewcommand{\arraystretch}{1.0}
\begin{table}[!h]
	\centering
	\caption{Regression result of the framework for turbidity estimation. The  turbidity is measured in Formazin Turbidity Unit (FTU).}
	\begin{tabular}{lSSSS}
		\toprule
		\multirow{ 3}{*}{Model} &\multicolumn{2}{c}{raw bands} &\multicolumn{2}{c}{with PCA} \\
		\cmidrule(ll){2-3}
		\cmidrule(ll){4-5}
		 & {$R^2$} & {RMSE}  & {$R^2$} & {RMSE} \\
		 & {in $\%$} & {in FTU} &  {in $\%$} &  {in FTU}  \\
		\midrule
        k-NN    & 70.0 & 0.3 & 88.8 & 0.2 \\
		RF      & 72.3 & 0.2 & 88.6 & 0.2 \\
		SVM     & 83.0 & 0.3 & 90.2 & 0.2 \\
		MARS    & 75.1 & 0.3 & 83.7 & 0.2 \\
		XGB     & 65.9 & 0.3 & 83.2 & 0.2  \\
		\bottomrule
	\end{tabular}
	\label{tab:results_turbidity}
\end{table}

\bibliographystyle{IEEEbib}
\bibliography{literature}

\end{document}